\documentclass{article}

\usepackage[utf8]{inputenc} 
\usepackage[T1]{fontenc}    
\usepackage{hyperref}       
\usepackage{url}            
\usepackage{booktabs}       
\usepackage{amsfonts}       
\usepackage{nicefrac}       
\usepackage{microtype}      
\usepackage{xcolor}         
\usepackage{multirow}
\usepackage{diagbox}
\usepackage{graphicx}
\usepackage{subfigure}
\usepackage{amsmath}
\usepackage{enumitem}
\usepackage{hyperref}

\usepackage[accepted]{icml2023}

\usepackage{amsmath}
\usepackage{amssymb}
\usepackage{mathtools}
\usepackage{amsthm}

\usepackage[capitalize,noabbrev]{cleveref}

\theoremstyle{plain}

\theoremstyle{definition}

\theoremstyle{remark}

\usepackage[textsize=tiny]{todonotes}

\icmltitlerunning{Neural Image Compression with Quantization Rectifier}

\begin{document}

\twocolumn[
\icmltitle{Neural Image Compression with Quantization Rectifier}




\begin{icmlauthorlist}
\icmlauthor{Wei Luo}{yyy}
\icmlauthor{Bo Chen}{zzz}
\end{icmlauthorlist}

\icmlaffiliation{yyy}{Department of Computer Science, Princeton University, New Jersey, United States}

\icmlaffiliation{zzz}{Department of Computer Science, University of Illinois at Urbana-Champaign, Illinois, United States}

\icmlcorrespondingauthor{Wei Luo}{wl4563@princetone.edu}
\icmlcorrespondingauthor{Bo Chen}{boc2@illinois.edu}

\icmlkeywords{Machine Learning, ICML}

\vskip 0.3in
]



\printAffiliationsAndNotice{}  

\begin{abstract}
Neural image compression has been shown to outperform traditional image codecs in terms of rate-distortion performance. However, quantization introduces errors in the compression process, which can degrade the quality of the compressed image. Existing approaches address the train-test mismatch problem incurred during quantization, the random impact of quantization on the expressiveness of image features is still unsolved. This paper presents a novel quantization rectifier (QR) method for image compression that leverages image feature correlation to mitigate the impact of quantization. Our method designs a neural network architecture that predicts unquantized features from the quantized ones, preserving feature expressiveness for better image reconstruction quality. We develop a soft-to-predictive training technique to integrate QR into existing neural image codecs. In evaluation, we integrate QR into state-of-the-art neural image codecs and compare enhanced models and baselines on the widely-used Kodak benchmark. The results show consistent coding efficiency improvement by QR with a negligible increase in the running time.
\end{abstract}

\section{Introduction}

Neural network (NN)-based image compression methods~\cite{balle2016end,balle2018variational,cheng2020image, minnenbt18} have shown superior coding efficiency to those of the conventional compression methods, such as BPG~\cite{bpg} and JPEG2000~\cite{jpeg2000}. 
Quantization discretizes image features by mapping continuous values to a limited set of discrete values for entropy coding, compressing the image~\cite{huffman1952method,witten1987arithmetic}. While current quantization methods address train-test mismatch, the random effects on feature expressiveness remain unresolved. Quantization uniformly maps continuous values to a single discrete value, introducing different degrees of noise depending on feature variability. It also unpredictably alters feature expressiveness. For instance, the quantization of features from the range of $[-0.5,0.5)$ to zero introduces noises in the range of $(-0.5,0.5]$. More importantly, quantization alters the expressiveness of the latent features in an unpredictable way.
In this paper, we propose a novel quantization rectifier (QR) that leverages spatial correlation in images to mitigate the impact of quantization. Specifically, we design a neural network architecture that predicts unquantized features from the quantized ones. To seamlessly integrate QR into a neural image codec, we introduce a soft-to-predictive (STP) training method. Here, we first softly train the original image compression model end-to-end until convergence. Then, we freeze the encoder network with hard quantization and optimize the decoder network, along with the QR network. QR bridges the gap between original and quantized features, preserving feature expressiveness for improved image reconstruction quality.
For evaluation, we incorporate our method into state-of-the-art neural network-based compression methods~\cite{balle2016end,balle2018variational,cheng2020image, minnenbt18}. We consistently improve all baseline models by up to 0.21 dB (PSNR) and 0.25 dB (MS-SSIM) without affecting the bitrate. QR is lightweight, with a minimal increase (0.7-5.4\%) in running time for most baselines.The contributions of this study are summarized as follows:
\begin{itemize}
\item We propose QR, a method that corrects quantized image features through prediction, preserving feature expressiveness and improving coding efficiency.
\item We develop the STP training procedure and a hyper-parameter exploration algorithm, enabling seamless integration of QR with existing neural image codecs.
\item We extensively evaluate QR on state-of-the-art neural image codecs, which demonstrate the superiority of QR consistently.
\end{itemize}

\section{Related Works}
\label{sec:related_works}

Quantization plays a vital role in image compression, enabling efficient storage and entropy coding~\cite{huffman1952method,witten1987arithmetic}. Recent advancements, including Additive Uniform Noises~\cite{balle2016end,balle2018variational} and Straight-Through Estimator~\cite{mentzer2018conditional, theis2017lossy, yin2019understanding} aim to tackle the train-test discrepancy arising from quantization. Soft-to-hard annealing (SA) approximates quantization using a differentiable function resembling hard rounding, but its training is fragile, requiring empirical determination of the annealing function. The Soft-Then-Hard (STH) strategy~\cite{guo2021soft} first learns a soft latent space and then resolves the train-test mismatch with hard quantization, partially addressing the issue. While these approaches address the train-test discrepancy, the expressiveness of latent features is still unpredictably affected by quantization. Our proposed approach effectively mitigates the impact of quantization on feature expressiveness and can be easily integrated into these techniques.

\section{Proposed Method}

\subsection{Formulation of Learned Compression Models}
\label{sec:background}

According to recent works~\cite{balle2018variational, cheng2020image, gao2022flexible}, the general procedure of neural image compression can be formulated as follows:
\begin{align}
    \label{eq:y} y &= g_{a}(x; \phi)\\
    \label{eq:yhat} \hat{y} &= Q(y)\\
    \label{eq:xhat} \hat{x} &= g_{s}(\hat{y}; \theta),
\end{align}
where $x$, $\hat{x}$, $y$, and $\hat{y}$ are the raw image, the reconstructed image, the latent feature before quantization, and the quantized latent feature, respectively. $\phi$ and $\theta$ are parameters of the encoder and decoder. In the encoding process $g_{a}$, latent feature $y$ is produced from the raw image $x$ (Eq.~\ref{eq:y}). For the quantization step $Q$, latent feature $y$ is quantized (rounded) to $\hat{y}$ (Eq.~\ref{eq:yhat}). During the decoding process $g_{s}$, quantized $\hat{y}$ is fed into the decoder network to obtain the reconstructed image $\hat{x}$ (Eq.~\ref{eq:xhat}). Since the quantization operation $Q$ is not differentiable in training, the quantization $Q$ is typically approximated by adding a uniform noise $\mathcal{U}(-0.5, 0.5)$ to the input. We define a probability model $p(\hat{y};\phi)$, which is parameterized by $\phi$ to compute the probability mass function (PMF) of quantized feature $\hat{y}$ as shown in Eq.~\ref{eq:pmf}.
\begin{align}
    \label{eq:pmf}
    p(\hat{y};\phi) &= \prod_i \int^{\hat{y}^{i} + 0.5}_{\hat{y}^{i} - 0.5} p(\hat{y}^{i};\phi)d\hat{y}^{i} \nonumber \\
    &= \prod_i \left(F\left(\hat{y}^{i} + 0.5;\phi\right) - F\left(\hat{y}^{i} - 0.5;\phi\right)\right),
\end{align}
where $F(\cdot;\phi)$ is the cumulative distribution function of $p(\cdot;\phi)$ and $i$ iterates over all symbols in $\hat{y}$. The goal of the image compression task is to minimize the rate-distortion loss function as shown in Eq.~\ref{eq:loss} with respect to parameters $\theta$ and $\phi$.
\begin{align}
    \label{eq:loss}
    \mathcal{L}_{\theta, \phi} &= \mathcal{R} (\hat{y}) + \lambda \mathcal{D} (x, \hat{x}) \\
    &= \mathbb{E} \underbrace{[-\log_{2}p(\hat{y};\phi)]}_{\text{rate}} + \underbrace{ \mathcal{D} (x, \hat{x})}_{\text{distortion}},
\end{align}
where the number of bits required to encode quantized $\hat{y}$ is represented by $\mathcal{R}=\mathcal{R} (\hat{y})$. The distortion between reconstruction image $\hat{x}$ and the original image $x$ is calculated by $\mathcal{D} (x, \hat{x})$, which is commonly evaluated by the peak signal-to-noise ratio (PSNR) or multiscale structural similarity (MS-SSIM)~\cite{wang2003multiscale} of the raw and reconstructed images. The encoder and entropy model parameter $\phi$, and the decoder parameter $\theta$ are jointly optimized to reduce the rate-distortion cost $\mathcal{R} + \lambda \mathcal{D}$, where $\lambda$ controls the rate-distortion trade-off.

The hyperprior adopted in many existing works~\cite{balle2018variational,mentzer2018conditional} is omitted from the above formulation for simplicity. However, such simplification does not affect the generality of our approach.

\subsection{Quantization Rectifier}
\label{sec:quantization_rectifier}

\begin{figure}[!htb]
     \centering
     \includegraphics[width=1\linewidth]{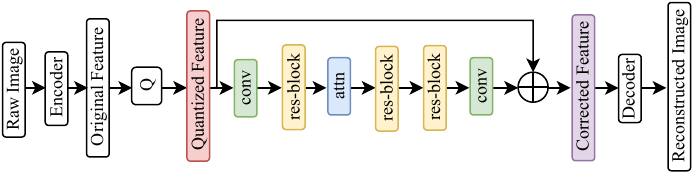}
     \caption{Quantization Rectifier Architecture.}
     \label{fig:quantization_rectifier}
\end{figure}

{\bf Definition.}
To mitigate the random impact of quantization, we introduce the quantization rectifier (QR). We take $y$ from Eq.~\ref{eq:y} as the input and train the corrected feature through a QR network, which is used to predict unquantized features from quantized ones. The effectiveness of the QR relies on its network design. The insight is to exploit the spatial correlation within the image feature that recovers itself even under noise. Inspired by the success of Model Diffusion~\cite{ho2020denoising} in image denoising, we design the QR network as shown in Fig.~\ref{fig:quantization_rectifier}. The QR network stays between the quantization step and the decoder. It consists of convolutional layers (conv), residual blocks (res-block), and attention layers (attn) that spatially correlate quantized features $\hat{y}$. We then add the output of the last conv to the quantized feature $\hat{y}$ and acquire the corrected feature $\Tilde{y}$ (Eq.~\ref{eq:tilde_y}). Next, $\Tilde{y}$ replaces $\hat{y}$ in the decoding phase as shown in Eq.~\ref{eq:tilde_x}.
\begin{align}
    \label{eq:tilde_y}\Tilde{y} &= \text{QR}(\hat{y})\\
    \label{eq:tilde_x}\hat{x} &= g_{s}(\Tilde{y}; \theta).
\end{align}
A more detailed architecture of the QR is described in Appendix~\ref{apx:quantization_rectifier_detailed_arch}. Compared to the network in Model Diffusion~\cite{ho2020denoising}, our network is configured with fewer layers for efficiency while being effective.
The QR is a versatile module, which can be seamlessly integrated into any neural image compression method that can be broken down into an encoder, a quantization module, and a decoder. There is no need to make significant modifications to the encoder and decoder components of the original image compression model.

{\bf Learning Rectifier.}
To facilitate the learning of the QR, we replace the loss function in Eq.~\ref{eq:loss} with Eq.~\ref{eq:newloss} that adds a \emph{feature distance} term to the original formulation.
\begin{equation}
    \label{eq:newloss}
    \mathcal{L}_{\theta,\phi,\psi} = \underbrace{\mathcal{R} (\hat{y})}_{\text{rate}} + \lambda  \underbrace{ \mathcal{D} (x, \hat{x})}_{\text{distortion}} + \alpha \underbrace{\mathcal{D}^f(y, \Tilde{y})}_{\text{feature distance}},
\end{equation}
where $\alpha$ is the learning coefficient controlling the relative learning rate of QR compared to that of the rate and distortion. An algorithm describing the exploration of $\alpha$ will be further detailed in Sec.~\ref{sec:training}. $\mathcal{D}^f(y, \Tilde{y})$ is a measure of the distance between the original and quantized image features, i.e., {feature distance}. The design of the feature distance is crucial to the learning of QR, where a smaller distance should reflect better preservation of feature expressiveness. We consider four commonly used distance terms: L1 distance~\cite{lasserre2009least}, L2 distance~\cite{park1990fast}, Smooth L1 distance~\cite{girshick2015fast}, and cosine similarity~\cite{salton1986introduction,deza2009cosine,manning2008introduction,ramos2003using}. Our empirical study shows that minimizing the L2 distance yields the best image quality with QR. Other measures like the L1 distance do not provide as promising results. Hence, we formulate the feature distance as in Eq.~\ref{eq:feature_distance}.
\begin{equation}
    \label{eq:feature_distance}
    \mathcal{D}^f(y,\Tilde{y}) = \lVert \mathbf{{y} - \Tilde{y}} \rVert_2.
\end{equation}

\subsection{Soft-to-predictive Training}
\label{sec:training}

Training is an essential step that integrates the QR into the neural image codec. We find the straightforward end-to-end training is sub-optimal due to the inter-dependency of the training of the codec and the QR. First, the learning of the QR relies on the stability of its input, which is the latent feature. Second, the stability of the latent features is contingent upon the convergence of the prediction network. Even a slight perturbation in the latent feature would disrupt the training process of the prediction network. Consequently, the disturbed prediction network would further affect the stability of the latent features. In such a vicious cycle, where the latent feature and prediction network constantly fail to converge, the overall training process is sub-optimal.

To address the sub-optimal training issue, we develop a soft-to-predictive (STP) training technique consisting of the soft and predictive training phases. In the soft training phase, the image is reconstructed based on Eq.~\ref{eq:y}, Eq.~\ref{eq:yhat}, and Eq.~\ref{eq:xhat}. Meanwhile, we learn parameters of the encoder ($\theta$), the decoder ($\phi$), and QR ($\psi$) based on the loss function Eq.~\ref{eq:newloss} softly with additive uniform noise. Although we do not apply Eq.~\ref{eq:tilde_y} and Eq.~\ref{eq:tilde_x} in the generation of $\hat{x}$, QR will still be learned to predict the feature, which warms up the next phase. 
In the predictive training phase, the image is reconstructed based on Eq.~\ref{eq:y}, Eq.~\ref{eq:yhat}, Eq.~\ref{eq:tilde_y}, and Eq.~\ref{eq:tilde_x}. The encoder $\theta$ is fixed with its output being hard quantized while the decoder ($\phi$) and QR ($\psi$) are optimized according to the loss function in Eq.~\ref{eq:newloss2}.
\begin{equation}
    \label{eq:newloss2}
    \mathcal{L}_{\theta,\psi} = \mathcal{D} (x, \hat{x}) + \alpha \mathcal{D}^f (y, \Tilde{y}),
\end{equation}
where only $\theta$ and $\psi$ are optimized. The bitrate $\mathcal{R}$ is omitted in Eq.~\ref{eq:newloss2} as its parameters are no longer optimized. $\lambda$ is also dropped in Eq.~\ref{eq:newloss2} for simplicity. In the predictive training phase, the latent feature and bitrate stay fixed, which stabilizes the training of QR.

During the predictive training phase, choosing the rectifier learning coefficient $\alpha$ in Eq.~\ref{eq:newloss} is non-trivial as its optimal value varies across different models and compression quality. Appendix~\ref{apx:learning_coefficient_exploration} demonstrate the learning coefficient exploration algorithm we proposed and its results.

\section{Experiments}
\label{sec:experiments}
\subsection{Experimental Setup}
\label{sec:setup}
To demonstrate the effectiveness of the QR, we apply it to four baseline neural image compression models: Factorized Prior~\cite{balle2016end}, Scale Hyperprior~\cite{balle2018variational}, Joint Hyperprior~\cite{minnenbt18}, and Attention-based Joint Hyperprior~\cite{cheng2020image}, denoted by ``Factorized'', ``Scale'', ``Joint'', and ``Attn'', respectively. The selected baseline models capture dominant neural image compression architectures. Accordingly, the models enhanced by the QR are represented by ``Factorized+QR'', ``Scale+QR'', ``Joint+QR'', and ``Attn+QR'', respectively.
According to CompressAI~\cite{compressai}, the four baseline models are previously trained on $256 \times 256$ image patches randomly extracted and cropped from the Vimeo90K dataset~\cite{xue2019video} with a batch size of $32$ using the Adam optimizer~\cite{kingma2014adam}. The test is performed on the commonly used Kodak image dataset~\cite{kodak_dataset}. We compare the performance of the four enhanced models against the corresponding baseline models in terms of rate and distortion trade-offs. Following many existing works~\cite{balle2016end,balle2018variational}, the rate is measured by bits per pixel (bpp) while the distortion is measured by either PSNR or MS-SSIM. MS-SSIM is converted to decibels $(-10 \log_{10} (1 -$ MS-SSIM$))$ to illustrate the difference clearly. For fairness, both baseline and enhanced models are optimized with MSE or MS-SSIM, depending on the distortion metric (PSNR or MS-SSIM). An description of the training configuration and a more detailed illustration of metrics and is in Appendix~\ref{apx:setups}.

\subsection{Coding Efficiency Improvement}
\label{sec:coding_efficiency_improvement}

\begin{figure}
\centering
\begin{subfigure}[PSNR]
{\includegraphics[width=.9\linewidth] {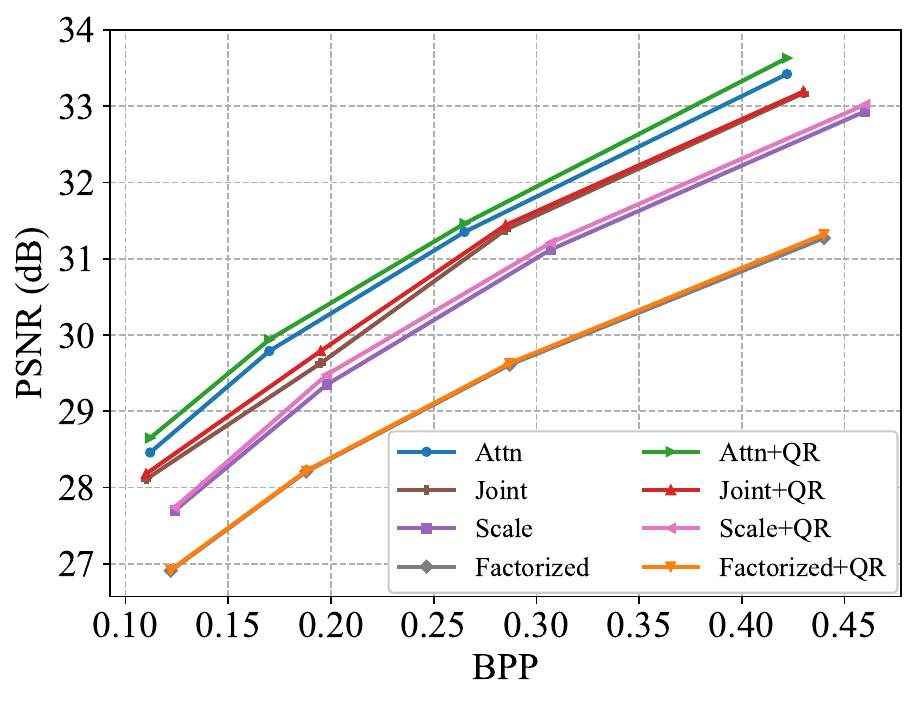}
  \label{fig:mse_rdtraeoff}
 }%
\end{subfigure}
\begin{subfigure}[MS-SSIM]
{\includegraphics[width=.9\linewidth] {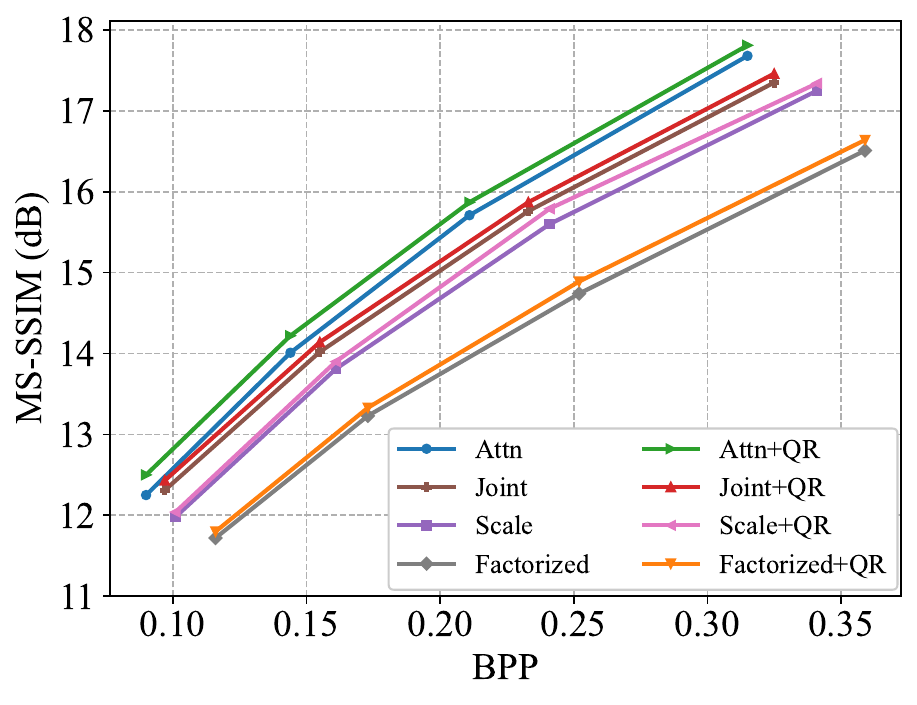}
  \label{fig:msssim_rdtraeoff}
 }%
\end{subfigure}
  \caption{Coding efficiency of baseline models and their enhanced versions by QR in terms of PSNR and MS-SSIM.}
  \label{fig:rdtraeoff}
\end{figure}

Fig.~\ref{fig:rdtraeoff} compares the coding efficiency of the baseline models without and with the proposed QR. Every point on curves in Fig.~\ref{fig:rdtraeoff} represents the bpp and distortion (PSNR or MS-SSIM) averaged over the Kodak image dataset~\cite{kodak_dataset} different compression quality levels $q \in \{1, 2, 3, 4\}$. For a specific baseline model at any given quality level, the average bpp value remains the same after applying QR. The reason is that the soft training process for the encoder with QR is identical to the training of the baseline model. After soft training, the encoder is fixed, so the average bpp value would not change. Comparing the baseline models to their corresponding enhanced versions in Fig.~\ref{fig:rdtraeoff}, we notice QR consistently improves all baseline models at various compression qualities in terms of both PSNR and MS-SSIM. Further, for a relatively more complex model, e.g., Attn, 
QR shows improvement by a wider margin than that of a simple model like Factorized. We speculate that a more complex model, with more parameters, can better leverage the reduced effect of quantization towards better image quality. Moreover, the improvement by QR is more evident utilizing MS-SSIM than utilizing PSNR.

The image quality of all models utilizing our proposed QR method surpasses that of the baseline models in both PSNR and MS-SSIM. Among all enhanced models, Attn+QR demonstrates the most substantial enhancement in PSNR, with an average quality improvement of 0.17 dB and a maximum improvement of 0.21 dB. With MS-SSIM, Attn+QR is still the best-performing one with a 0.19 dB average and 0.25 dB maximum improvement over Attn. While Factorized+QR exhibits a relatively smaller improvement compared to the other enhanced models in PSNR, its improvement is significant in MS-SSIM, with an average of 0.12 dB and a maximum of 0.15 dB. The numerical results are summarized in Tab.~\ref{tab:psnrsim_improvement} shown in Appendix~\ref{apx:psnrsim_table}.

A detailed evaluations regarding quantization error reduction of the proposed QR component can be found in Appendix~\ref{apx:quantization_error_reduction}.

\subsection{Processing Speed} 
\begin{table}[htb!]
  \centering
  \caption{Runtime cost increase in milliseconds after applying the quantization rectifier, evaluated on NVIDIA RTX 2080 Ti GPU (2080) and NVIDIA RTX 3090 Ti GPU (3090).}
  \label{tab:runtime_cost}
  \begin{tabular} {c |c|c|c|c}
  \toprule
{HW}&{Attn}&{Joint}&{Scale}&{Factorized}\\
\midrule
2080&5.2\%&0.7\%&4.6\%&16.6\%\\ 
3090&5.4\%&0.8\%&5.4\%&17.1\%\\ 
\bottomrule
  \end{tabular}
\end{table}

In Tab.~\ref{tab:runtime_cost}, we compare the average processing time per frame of the baseline models and their enhanced versions by QR, including encoding and decoding, on the Kodak dataset. During the time measurement, we factor out the time spent in the conversion between symbol likelihoods and bits to precisely show the impact of QR on the neural network-related computation. Tests are performed on NVIDIA RTX 2080 Ti and NVIDIA RTX 3090 Ti GPUs. Our method slightly increases processing time by 0.7-5.4\% for most baselines (Attn, Joint, and Scale), while Factorized is more affected as its processing time is already short due to its simplest network architecture.

\section{Conclusion}

We introduce a Quantization Rectifier (QR) method to enhance neural image compression. QR utilizes spatial correlation in images to predict features before quantization, preserving their expressiveness. Our method includes a soft-to-predictive training approach that allows seamless and optimal integration of QR into existing neural image codecs. Experimental results consistently demonstrate the effectiveness of QR across various state-of-the-art neural image codecs.

\bibliography{example_paper}
\bibliographystyle{icml2023}

\newpage
\appendix
\onecolumn
\section{Detailed Quantization Rectifier Architecture}
\label{apx:quantization_rectifier_detailed_arch}

\begin{figure}[!htb]
\centering
\includegraphics[width=0.85\linewidth]{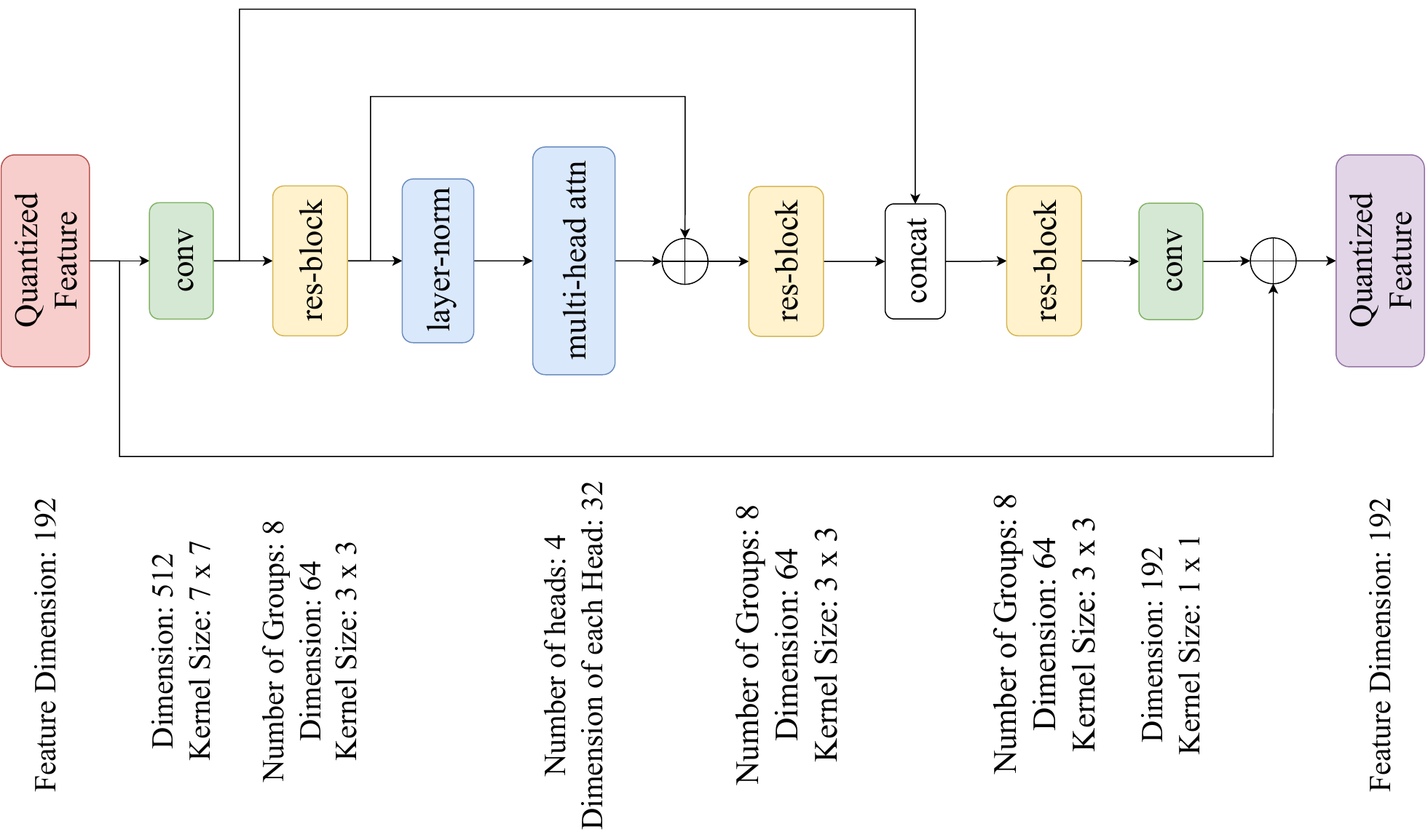}
\caption{Detailed Architecture of the Quantization Rectifier.}
\label{fig:appendix_arch}
\end{figure}

In accordance with Sec.~\ref{sec:quantization_rectifier}, the architecture presented in Fig.~\ref{fig:appendix_arch} provides a comprehensive illustration of the Quantization Rectifier (QR) network. For the purpose of clarity, the encoder and decoder components depicted in Fig.~\ref{fig:quantization_rectifier} are omitted in this particular representation. The architecture consists of convolutional layers (conv), which are intertwined with groups of residual blocks (res-block) and multi-head attention layer (attn). These attention layers spatially correlate the quantized features $\hat{y}$.

Initially, the quantized feature with a dimension of $192$ is inputted into a conv with a dimension of $512$, employing a kernel size of $7 \times 7$ and a padding of $3$. The output of this conv is then directed to the first set of grouped res-block, consisting of eight groups, with each group having a dimension of $64$ and a kernel size of $3 \times 3$. The resulting output from the res-block is normalized through layer normalization (layer-norm) and serves as the input for a multi-head attn with four heads, where each head possesses a dimension of $32$. The output of this attn is subsequently added to the output of the first res-block layer. This summation then serves as the input for the second set of res-block groups, which mirror the architecture of the first group. The output of this second set of res-block is concatenated with the output of the initial conv and is then fed into the final group of res-block. The concatenated output undergoes a final conv with a dimension of $192$ and a kernel size of $1 \times 1$. Lastly, the output from this conv is added to the quantized feature $\hat{y}$, resulting in the corrected feature $\Tilde{y}$, which retains the same dimension as the original quantized feature $\hat{y}$.

\section{Learning Coeeficient Exploration in Training}
\label{apx:learning_coefficient_exploration}

\begin{figure}
  \centering
  \includegraphics[width=0.4\linewidth]{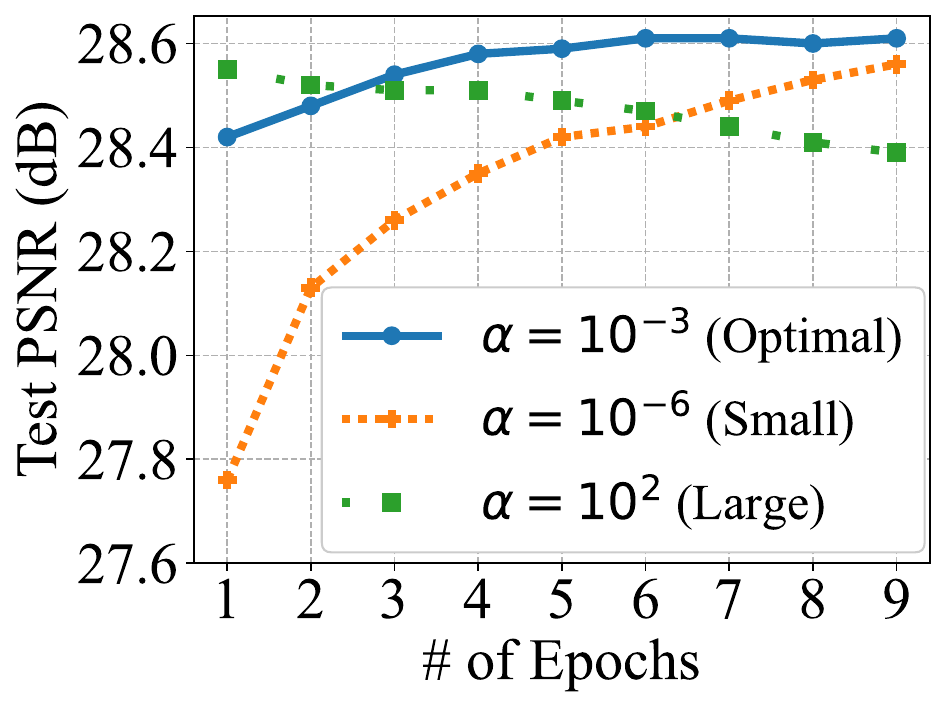}
  \caption{The impact of rectifier learning coefficient on image quality.}
  \label{fig:alpha_impact}
\end{figure}

{\bf Rectifier Learning Coefficient Exploration Algorithm.}
During the predictive training phase as mentioned in Sec.~\ref{sec:training}, choosing the rectifier learning coefficient $\alpha$ in Eq.~\ref{eq:newloss} is non-trivial as its optimal value varies across different models and compression quality. Fig.~\ref{fig:alpha_impact} compares the image quality tested on the Kodak dataset, resulting from different rectifier learning coefficients. 
An optimal coefficient, e.g., $10^{-3}$, allows the image quality to start at a high value and converge in a few epochs, e.g., 6 epochs in Fig.~\ref{fig:alpha_impact}. If the learning coefficient is too small, e.g., $10^{-6}$, it will take a long time for the model to converge at a sufficiently good image quality. Conversely, if the learning coefficient is too large, e.g., $10^2$, the rectifier changes too fast for the decoder to converge, which degrades image quality. To tackle this issue, we introduce a rectifier learning coefficient exploration method that automatically finds the optimal learning coefficient for different models and compression quality.

One of our key findings is, there exists an optimal learning coefficient where increasing or decreasing it would only monotonically degrade coding efficiency. Based on this finding, we describe the exploration strategy for a specific model and compression quality as follows: i) start the exploration at an initial learning coefficient $\alpha=\alpha_{max}$, ii) train the codec as specified in Sec.~\ref{sec:training} using $\alpha$ until the loss (Eq.~\ref{eq:newloss}) stops improving for $M$ consecutive epochs, iii) multiply the learning coefficient $\alpha$ by 0.1, iv) continue step ii) if the learning coefficient is no smaller than a pre-defined lowest learning coefficient $\alpha_{min}$ or stop exploration otherwise. $M$ is set to $3$, which confidently finds the non-improving loss. Considering the efficiency of exploration, we adopt a relatively small dataset in exploration, which remains as effective as the big one in predictive training.

\begin{table*}[htb!]
  \centering
  \caption{Rectifier Learning Coefficient Exploration: distortion measured in PSNR (dB) at compression qualities $q \in {1,2,3,4}$ using Flickr image dataset.}
  \label{tab:pe}
  \begin{tabular} {c| c c c c |c c c c}
  \toprule
\multirow{2}{*}{{Coefficient}}&\multicolumn{4}{c|}{Attn}&\multicolumn{4}{c}{Joint}\\ \cmidrule{2-9}
&$1$&$2$&$3$&$4$&$1$&$2$&$3$&$4$\\
\midrule
$10^{-1}$&28.56&29.85&31.37&33.47&                          28.15&29.71&31.35&33.11\\ 
$10^{-2}$&28.58&29.87&31.39&33.49&                          28.16&29.73&31.38&33.12\\ 
$10^{-3}$&{\bf 28.61}&{\bf 29.88}&{\bf 31.41}&{\bf 33.52}&  28.16&29.73&{\bf31.39}&{\bf 33.13}\\ 
$10^{-4}$&28.59&29.86&31.40&33.50&                          {\bf28.17}&{\bf29.74}&31.38&33.12\\ 
$10^{-5}$&28.58&29.84&31.38&33.49&                          28.16&29.74&31.37&33.11\\ 
\midrule
\multirow{2}{*}{{Coefficient}}&\multicolumn{4}{c|}{Scale}&\multicolumn{4}{c}{Factorized}\\ \cmidrule{2-9}
&$1$&$2$&$3$&$4$&$1$&$2$&$3$&$4$\\
\midrule
$10^{-1}$&27.63&29.35&31.09&32.93&                  26.90&28.19&29.60&31.26\\ 
$10^{-2}$&27.68&29.40&31.13&{\bf32.96}&             {\bf 26.91}&{\bf 28.21}&29.60&31.27\\ 
$10^{-3}$&27.70&29.42&{\bf31.16}&32.93&             26.91&28.20&{\bf 29.61}&31.28\\ 
$10^{-4}$&{\bf27.73}&{\bf29.44}&31.15&32.92&        26.90&28.20&29.59&{\bf 31.28}\\ 
$10^{-5}$&27.70&29.43&31.14&32.92&                  26.90&28.19&29.58&31.27\\ 
\bottomrule
  \end{tabular}
\end{table*}

{\bf Rectifier Learning Coefficient Exploration Result.}
Choosing the rectifier learning coefficients $\alpha$ in Eq.~\ref{eq:newloss} is critical to the performance of the compression model. We empirically set an $\alpha_{max}=10^{-1}$ and $\alpha_{min}=10^{-5}$ for PSNR to cover feasible values of the learning coefficient. We explore learning coefficients between $\alpha_{max}$ and $\alpha_{min}$ following the exploration strategy in Sec.~\ref{sec:training}. The exploration for each coefficient value usually completes with 15 rounds of iterations. Tab.~\ref{tab:pe} shows the PSNR values for all models achieved at explored coefficients, where the coefficient producing the best PSNR (highlighted) is selected for predictive training. It is also observed that the PSNR performance monotonically degrades when the coefficient increases above or decreases below the selected value. Note that, as we round off the PSNR value to two decimal places, some adjacent rows with different coefficients may show the same PSNR value, e.g., 26.91 dB at $q=1$ for coefficients $10^{-2}$ and $10^{-3}$ of Factorized. However, only the highlighted PSNR value is higher than others in the same column.
The exploration of the rectifier learning coefficient efficiently identifies the optimal coefficient that ensures a better predictive training result than that without exploration.

Note that we do not repeat the same exploration process for MS-SSIM as we do for PSNR. The reason is we empirically find the distribution of optimal coefficients is similar for PSNR and MS-SSIM on different models, except that the optimal coefficient for MS-SSIM is roughly $10\times$ smaller than that for PSNR. As a result, we multiply the optimal coefficient in Tab.~\ref{tab:pe} by $0.1$ for MS-SSIM during predictive training phase.

\section{Training Configuration and Detailed Experimental Metrics}
\label{apx:setups}

{\bf Training configuration.}
For enhanced models, the rectifier learning coefficient exploration and the STP training phases are conducted on the Flickr image dataset~\cite{flickr_dataset} and the ImageNet dataset~\cite{5206848}, respectively. Similar to the baseline training, the exploration and STP training phases adopt a batch size of $32$. We fix the learning rate at $10^{-6}$ in both phases, which yields the best performance. The training is performed on a desktop with 2 NVIDIA RTX 3090 GPUs.

{\bf Metrics.}
As mentione in~\ref{sec:setup}, we compare the performance in terms of rate and distortion trade-offs, measured by bpp and PSNR or MS-SSIM. To show the performance of our approach at different compression qualities, we repeat our experiments at compression quality levels $q \in \{1, 2, 3, 4\}$ of the codec models, where a greater value of $q$ corresponds to a greater value of $\lambda$ in Eq.~\ref{eq:loss}.
To demonstrate the capability of our approach in preserving the expressiveness of image features, we introduce a novel metric, the quantization error $\epsilon_Q$. The quantization error is defined as the L2 distance  between the input to the quantization operation ($y$) and the input to the decoder, which is either the result of quantization ($\hat{y}$) or QR ($\Tilde{y}$) depending on whether a model is enhanced by QR, as shown in Eq.~\ref{eq:qe}.
\begin{equation}\label{eq:qe}
\epsilon_Q =
\begin{cases}
    \lVert {\Tilde{y} - y} \rVert_2 & \text{if using QR} \\
    \lVert {\hat{y} - y} \rVert_2 & \text{if not using QR. } 
\end{cases}
\end{equation}

\section{Numerical Results of Coding Efficiency Improvement}
\label{apx:psnrsim_table}

As mentioned in Sec.~\ref{sec:coding_efficiency_improvement}, Tab.~\ref{tab:psnrsim_improvement} statistically shows the benefits of QR in terms of PSNR and MS-SSIM regarding the average and maximum values over all compression qualities.

\begin{table*}[htb!]
  \centering
  \caption{Image quality improvement in PSNR and MS-SSIM.}
  \label{tab:psnrsim_improvement}
  \begin{tabular} {c |c |c|c|c| c| c| c| c }
  \toprule
\multirow{2}{*}{Metrics (dB)}&\multicolumn{2}{c|}{Attn}&\multicolumn{2}{c|}{Joint}&\multicolumn{2}{c|}{Scale}&\multicolumn{2}{c}{Factorized}\\\cline{2-9}
&Avg&Max&Avg&Max&Avg&Max&Avg&Max\\
\midrule
$\uparrow$ PSNR&0.17$\pm$0.04& 0.21&0.08$\pm$0.05&0.16 &0.09$\pm$0.03&0.13&0.02$\pm$0.01&0.05\\ 
$\uparrow$ MS-SSIM&0.19$\pm$0.05& 0.25&0.12$\pm$0.01&0.12 &0.11$\pm$0.05&0.19&0.12$\pm$0.03&0.15\\ 
\bottomrule
  \end{tabular}
\end{table*}

\section{Evaluation of Quantization Error Reduction of QR}
\label{apx:quantization_error_reduction}

{\bf Quantization Error Reduction Result.}
\begin{figure}
\centering
\begin{subfigure}[Model optimized for PSNR]
{\includegraphics[width=.46\linewidth] {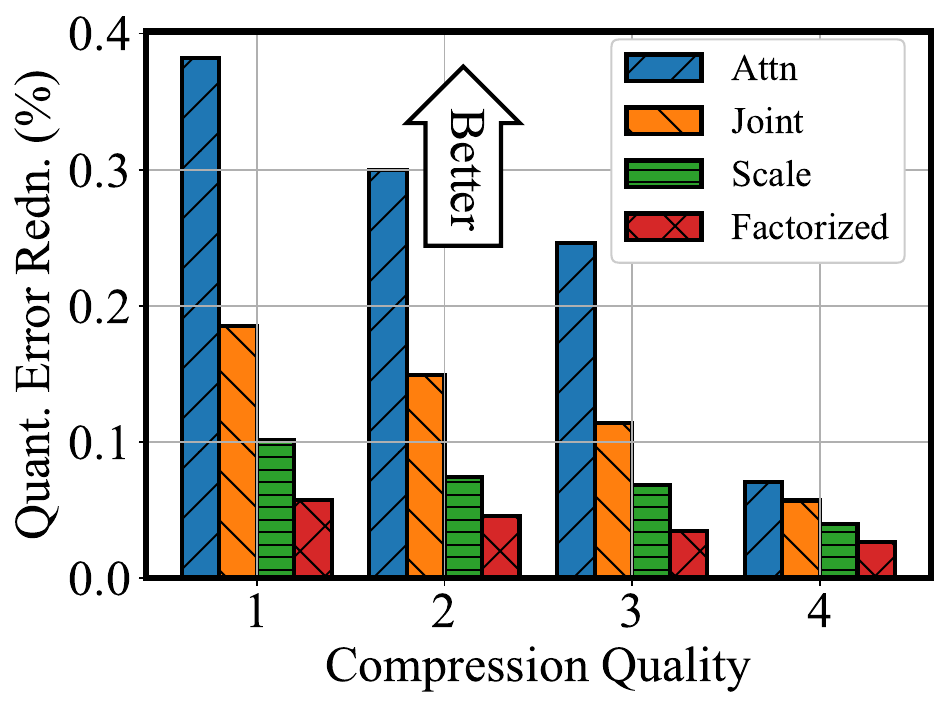}
  \label{fig:mse_norm}
 }%
\end{subfigure}
\begin{subfigure}[Model optimized for MS-SSIM]
{\includegraphics[width=.46\linewidth] {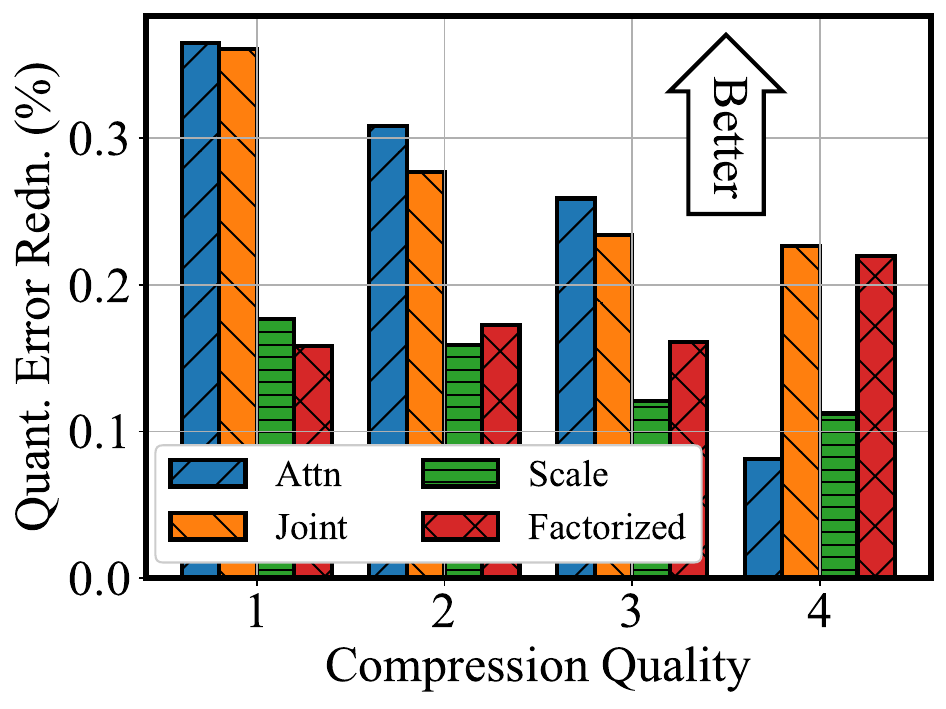}
  \label{fig:msssim_norm}
 }%
\end{subfigure}
  \caption{Quantization error reduction.}
  \label{fig:quant_err_redn}
\end{figure}
Fig.~\ref{fig:quant_err_redn} shows the reduction of quantization error by QR in percentage compared against all baseline models at various compression qualities. The reduction is generally more significant for more complex models like Attn and Joint. Meanwhile, a lower compression quality tends to magnify the reduction.
Notably, we observe the maximum quantization error reduction of 38\% and 36\% for Attn at compression quality $q=1$, when the model is optimized for PSNR and MS-SSIM, respectively. We also notice models optimized for MS-SSIM, e.g., Factorized, exhibit randomness that causes the reduction to be slightly improved with a higher compression quality.

\begin{figure}
\centering
\begin{subfigure}[Model optimized for PSNR]
{\includegraphics[width=.46\linewidth] {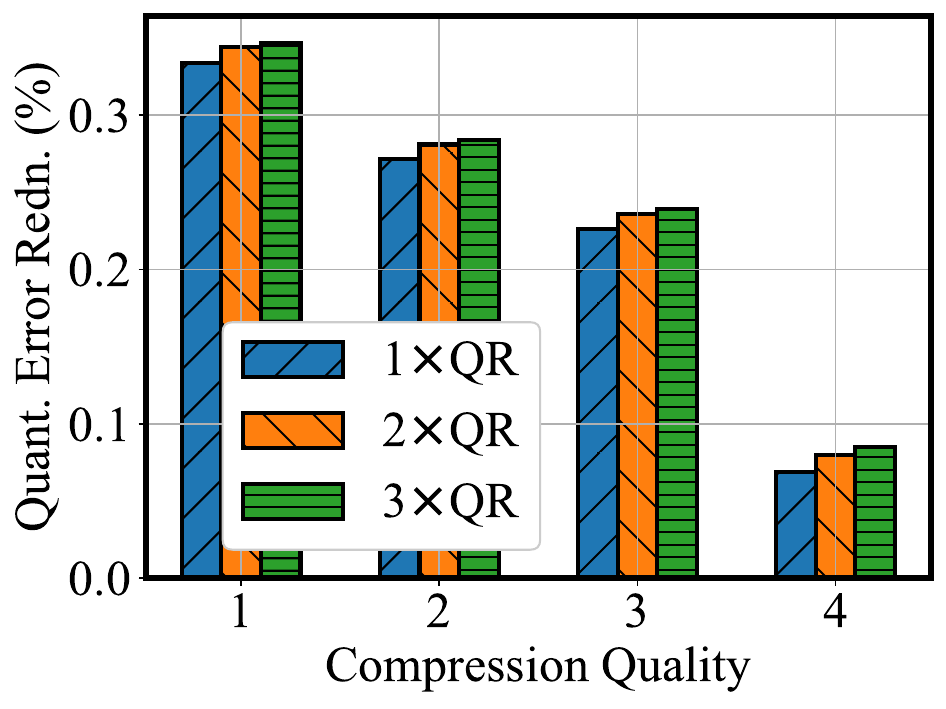}
  \label{fig:mse_norm_vs_layer}
 }%
\end{subfigure}
\begin{subfigure}[Model optimized for MS-SSIM]
{\includegraphics[width=.46\linewidth] {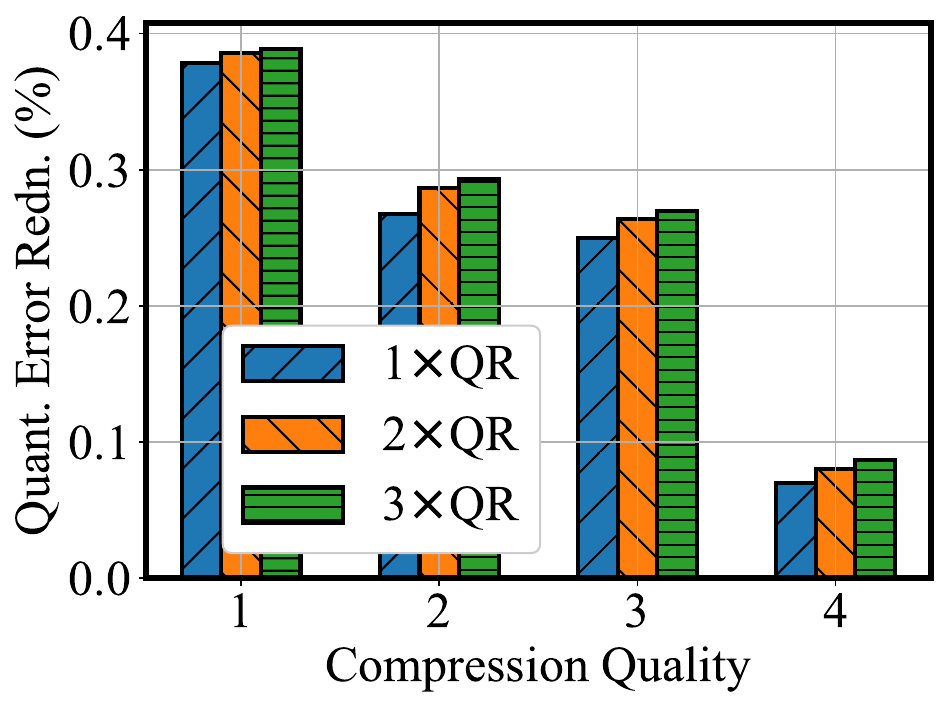}
  \label{fig:msssim_norm_vs_layer}
 }%
\end{subfigure}
  \caption{Impact of different numbers of quantization rectifiers on quantization error reduction.}
  \label{fig:norm_vs_layer}
\end{figure}

\begin{figure}
\centering
\begin{subfigure}[Model optimized for PSNR]
{\includegraphics[width=.46\linewidth] {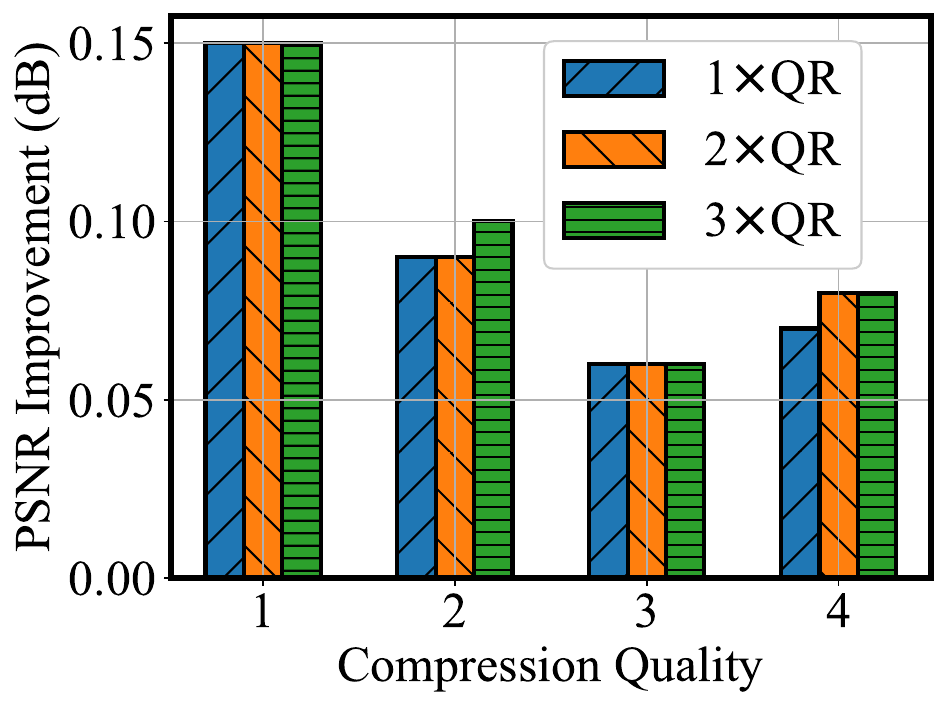}
  \label{fig:psnr_vs_layer}
 }%
\end{subfigure}
\begin{subfigure}[Model optimized for MS-SSIM]
{\includegraphics[width=.46\linewidth] {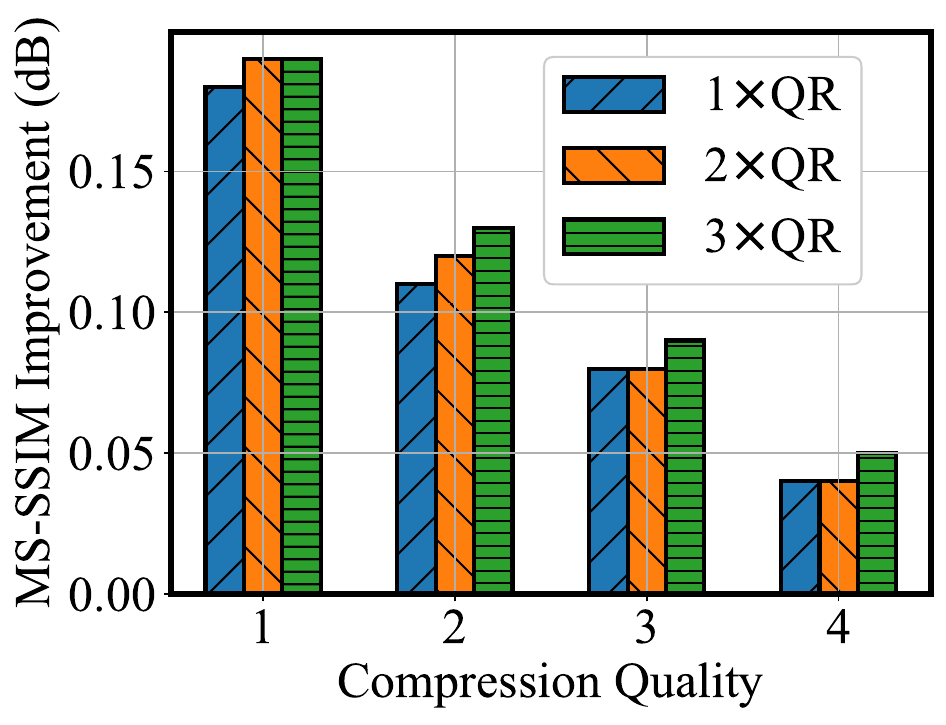}
  \label{fig:msssim_vs_layer}
 }%
\end{subfigure}
  \caption{Impact of different numbers of quantization rectifiers on image quality.}
  \label{fig:psnr_msssim_vs_layer}
\end{figure}

{\bf Analysis of Quantization Rectifier.} 
With the performance gain of one QR, a natural idea is to embed multiple QRs into a codec. To this end, we apply multiple sequentially connected QRs to a baseline codec and analyze the improvements in image quality and quantization error reduction. Specifically, we compare the performance of one, two, and three QR(s), denoted by $1\times$QR, $2\times$QR, and $3\times$QR, respectively. Attn is used as the baseline model for different compression qualities. As Fig.~\ref{fig:norm_vs_layer} shows, the quantization error reduction is already significant when we have one QR, which is up to 35\% for PSNR and 39\% for MS-SSIM models. Despite more QRs further reduce the quantization error, the benefit of an additional QR, being less than 2\%, is rather incremental in both PSNR and MS-SSIM. 
Similarly in Fig.~\ref{fig:psnr_msssim_vs_layer}, the improvement in the image quality measured by PSNR and MS-SSIM is significant with the first QR, which is up to 0.15 dB and 0.19 dB, respectively. After that, when we use two or three QRs, the image quality gain from one additional QR is at most 0.01 dB. 
When the compression quality changes, the above observations still apply.
Given the fact that the computation and memory (storage) overhead of QR linearly increases with its number, affecting the training and encoding/decoding of the codec, we do not pursue a network design with more than one QR.

\end{document}